\documentclass{svproc}
\usepackage{type1cm}        
\usepackage{graphicx}        

\usepackage[bottom]{footmisc}

\usepackage{newtxmath}       

\usepackage{url}

\usepackage[numbers]{natbib}
\usepackage{graphicx}
\usepackage{verbatim,color}
\usepackage{hyperref}
\usepackage{subcaption}

\newcommand*{\normally}{\mathrel{\ooalign{$\mid $\hfil\cr\kern+1pt$\thicksim$}}}
\newcommand*{\nnormally}{\mathrel{\ooalign{$\mid$\hfil\cr\kern+1pt$\thicksim$}\negthickspace \negthickspace /} } 

\def\F{\mathcal F}

\def\prev{\mathbb{P}}

\newcommand{\no}[1]{\overline{#1}}

\begin{document}
\sloppy
	\mainmatter              
	\title{Probabilistic Weak Contraposition}
	\titlerunning{Probabilistic Weak Contraposition}  
	%
	\author{Lydia Castronovo\inst{1} \and Niki Pfeifer\inst{2}
		 \and Giuseppe Sanfilippo\inst{1}}
	\authorrunning{Lydia Castronovo \and Niki Pfeifer
		 \and Giuseppe Sanfilippo} 
	%
	\tocauthor{Lydia Castronovo, Niki Pfeifer, and Giuseppe Sanfilippo}
	\institute{Department of Mathematics and Computer Science, University of Palermo, Italy\\
		\email{\{lydia.castronovo,giuseppe.sanfilippo\}@unipa.it},\\ 
		\and
		Ruhr University Bochum, 
        Department of Philosophy I, Germany\\
		\email{niki.pfeifer@rub.de}}
	
	\maketitle              
	
	\begin{abstract}
We study Weak Contraposition in a probabilistic framework. While Contraposition is not valid in nonmonotonic reasoning and probability logic, Weak Contraposition preserves nonmonotonicity and provides a more plausible inference pattern. Using a coherence-based approach to probability, we provide a probabilistic semantics for Weak Contraposition and derive probability propagation rules for the conclusion given the premises. We study both precise and interval-valued probability assessments and consider cases with and without background knowledge. We also extend the framework to iterated conditionals using conditional random quantities and coherent previsions.

		\keywords{Weak Contraposition, Coherence, Nonmonotonic Reasoning, Probabilistic Semantics, Iterated Conditionals}
	\end{abstract}
    \section{Introduction}

In this paper we investigate Weak Contraposition, which has been proposed as a nonmonotonic version of Contraposition (\cite{freund1991}) in a probabilistic setting.   Contraposition is the inference of \emph{if $\no{B}$, then $\no{A}$} from the single premise \emph{if $A$, then $B$}. It is classically logically valid (when a conditional \emph{if $A$, then $B$} is interpreted as the material conditional \emph{$\no{A} \vee B$}) but not valid in default reasoning (when the conditional is interpreted by the default \emph{if $A$ normally $B$}). 
Contraposition  can thus be problematic in non-classical logics.  Contraposition is also not valid in standard logics for counterfactual conditionals (\cite[p. 36]{lewis73}). 
Adding Contraposition to the nonmonotonic reasoning System P (\cite{kraus90}), for example, would render System~P monotonic,\footnote{Assume $A \normally C$, where $\normally$ is the nonmonotonic consequence relation. By Contraposition, $\no C \normally \no A$. From Right Weakening we obtain $\no{C} \normally (\no{A} \vee \no{B})$ from $\models\no{A} \rightarrow (\no{A} \vee \no{B})$ and $\no{C} \normally \no{A}$, where ``$\models$'' denotes logical entailment. By Contraposition again, $(A \land B) \normally C$, which makes System P monotonic.}  
while Weak Contraposition preserves the nonmonotonicity property (\cite{freund1991}).  
Weak Contraposition is  the inference of \emph{if $\no{B}$ and $H$, normally $\no{A}$} from the two premises \emph{if $A$ and $H$, normally $B$} and \emph{it is not the case that: if $H$, normally $B$}, where $H$ denotes some ``arbitrary background knowledge'' (\cite[p. 195]{freund1991}). 
To illustrate why Contraposition is invalid for everyday reasoning, consider the following inference: from the premise \emph{if a student studies hard ($A$), she will normally pass the exam ($B$)} the conclusion \emph{if she does not pass the exam ($\no{B}$),  she normally did not study hard ($\no{A}$)} does not necessarily follow. For example, it could be that the exam was exceptionally hard.

However, when we consider Weak Contraposition, we obtain the following intuitively valid inference: from the two premises \emph{if a student studies hard ($A$) and takes the exam ($H$), she will normally pass the exam ($B$)} and \emph{it is not the case that: if a student takes the exam ($H$), she will normally pass ($B$)} we can conclude that \emph{if she does not pass ($\no{B}$) and took the exam ($H$), she normally did not study hard ($\no{A}$)}. Here, the second premise suggests that it is not true that simply taking the exam is usually enough to pass the exam. This inference is also intuitively valid when $H$ is omitted.

Psychologically, there are only few empirical studies devoted to Contraposition:  they indicate that most people do not accept Contraposition as a valid inference scheme, neither for indicative conditionals  (\cite{pfeifer06c,pfeifer10a,PolitzerBaratgin2016}) nor for counterfactual conditionals (\cite{pfeifer15a}). Experimental studies on Weak Contraposition are a desideratum for future work.

From a probabilistic point of view, we recall that Contraposition is not p-valid (\cite{adams75}). From $P(B|A)=x$ it follows that any value of $P(\no{A}|\no{B})\in [0,1]$ is a coherent extension (i.e., $P(B|A)$ does not constrain $P(\no{A}|\no{B})$), where $B|A$ represents the conditional \emph{if $A$, then $B$} as a conditional event. Thus, in order to obtain a probabilistic informative inference rule, we need an additional premise. In particular, we give a probabilistic semantics of Weak Contraposition and prove probability propagation rules for precise and imprecise cases. \\
The paper is organized as follows.
In Section \ref{SEC:ProbPropWC} we investigate Weak Contraposition in a probabilistic framework. We derive probability propagation rules that characterize the coherent bounds for the probability of the conclusion given the premises, both in the presence and in the absence of background knowledge. The analysis is developed for precise probability assessments as well as for interval-valued assessments. In Section \ref{SEC:ITERATED} we extend the study to the case of iterated conditionals by adopting the framework of conditional random quantities and coherent previsions, showing that analogous propagation rules hold in this setting. Finally, Section \ref{SEC:ConcludRem} concludes with a discussion of the results and outlines directions for future research. 
Some proofs are omitted due to the lack of space. 
\section{Probability Propagation Rules for Weak Contraposition}
\label{SEC:ProbPropWC}
Weak Contraposition was proposed by \cite{freund1991} in the context of default reasoning (here, $H$ denotes a basic knowledge base): 
\begin{equation}\label{EQ:WCdefault}
(A\wedge H) \normally B,\;\; H \nnormally B \models (\no{B}\wedge H) \normally \no{A},
\end{equation}
where $A \normally C$ denotes the default \emph{if $A$ normally $C$} and $A \nnormally C$ denotes the negated default \emph{it is not the case that: if $A$ normally $C$}. Like in probability logic, Contraposition does not hold in default reasoning:  $A \normally B$ alone does not entail $\no{B} \normally \no{A}$.  In our semantics (see \cite{PS24connexivity,PfSa24}), we interpret the default $A \normally C$ and the negated default $ A \nnormally C$ as the probability constraints $P(C|A)=1$ and $P(C|A)<1$ (or equivalently $P(\no C|A)>0$), respectively. Then, the probabilistic version of Weak Contraposition with knowledge base $H$ is:
\begin{equation}\label{EQ:WCwithH}
P(B|A\wedge H)=1,\; P(\no B|H)>0 \Longrightarrow P(\no A|\no{B}\wedge H)=1.
\end{equation}

In particular, when $H$ coincides with the sure event $\Omega$, the probabilistic Weak Contraposition amounts to 
 \begin{equation}\label{EQ:WCtop}
     P(B|A)=1,\; P(\no B)>0 \Longrightarrow P(\no A|\no{B})=1. 
\end{equation}
In the next result we study the probability propagation rule of Weak Contraposition, which allows us to validate (\ref{EQ:WCtop}).

\begin{theorem}\label{THM:WC1}
 Given any coherent probability assessment $(x,y)\in [0,1]^2$ on the family $\{B|A, \no{B}\}$, with $A,B$ logically independent (and  hence with   $A,\no{B} \neq \emptyset$), the probability assessment $z = P(\no{A}|\no{B})$ is a coherent extension of $P(B|A)=x$ and $P(\no{B})=y$ if and only if  $z\in [z'(x,y),z''(x,y)]$, where 
 \begin{equation}
\label{EQ:LB&UB} 
 z'(x,y)=\left\{
 \begin{array}{ll}
 \frac{x+y-1}{xy}, & \text{ if } x+y>1,\\
 0, &\text{ if } x+y\leq 1 \, .
 \end{array}
 \right.\quad
\text{ and }\quad z''(x,y)=1.
\end{equation} 
\end{theorem}
\begin{proof}
     The proof is  omitted due to the lack of space.
\end{proof}
\begin{remark}\label{REM:InstantiationWC}
    By instantiating $x=1$ and $y>0$ in Theorem~\ref{THM:WC1}, it follows that $z'(1,y)=z''(1,y)=1$, which validates  Equation~(\ref{EQ:WCtop}). 
\end{remark}

From Theorem \ref{THM:WC1},  recalling that the set of all coherent assessments on $\{B|A,\no{B}\}$ coincides with the unit square $[0,1]^2$, it follows 
\begin{corollary}\label{COR:WC1}
Let $A$, $B$,  be any logically independent events. Then, the set $\Pi$ of all coherent probability assessments $(x,y,z)$ on  $\F=\{B|A, \no{B}, \no{A}|\no{B}\}$, is 
$
\Pi=\{(x,y,z): (x, y)\in[0,1]^2, z \in [z'(x,y), z''(x,y)]\},	
$
 where  $z'(x,y)$ and $z''(x,y)$ are  defined in (\ref{EQ:LB&UB}).
\end{corollary}
We recall that in case of logical dependencies or with some further coherent assessments, the set
of all coherent assessments may be smaller than that one given in Corollary \ref{COR:WC1}.

We now study the probability propagation rule of Weak Contraposition including the background knowledge $H$, which allows us to validate (\ref{EQ:WCwithH}).

\begin{theorem}\label{THM:WC}
	{\rm
 Given any coherent probability assessment $(x,y)\in [0,1]^2$ on the family $\{B|AH, \no{B}|H\}$, with $A,B,H$ logically independent (and hence with $A,\no{B},H \neq \emptyset$), the probability assessment $z = P(\no{A}|\no{B}H)$ is a coherent extension of $P(B|AH)=x$ and $P(\no{B}|H)=y$ if and only if  $z\in [z'(x,y),z''(x,y)]$, where 
 \begin{equation}
\label{EQ:LB&UBWITHH}
  z'(x,y)=\left\{
 \begin{array}{ll}
 \frac{x+y-1}{xy}, & \text{ if } x+y>1,\\
 0, &\text{ if } x+y\leq 1 \, .
 \end{array}
 \right.\quad
\text{ and }\quad z''(x,y)=1.
\end{equation}
}
\end{theorem}
\begin{proof}
 The proof is  omitted due to the lack of space.
\end{proof}
From Theorem \ref{THM:WC},  recalling that the set of all coherent assessments on $\{B|AH,\no{B}|H\}$ coincides with the unit square $[0,1]^2$, it follows 
\begin{corollary}\label{COR:WC}
Let $A$, $B$, and $H$ be any logically independent events. Then, the set $\Pi$ of all coherent probability assessments $(x,y,z)$ on  $\F=\{B|AH, \no{B}|H, \no{A}|\no{B}H\}$, is 
$
\Pi=\{(x,y,z): (x, y)\in[0,1]^2, z \in [z'(x,y), z''(x,y)]\},	
$
 where  $z'(x,y)$ and $z''(x,y)$ are  defined in (\ref{EQ:LB&UBWITHH}).
\end{corollary}

\begin{figure}
\vspace{-0.7cm}
 \centering    \includegraphics[width=1\linewidth]{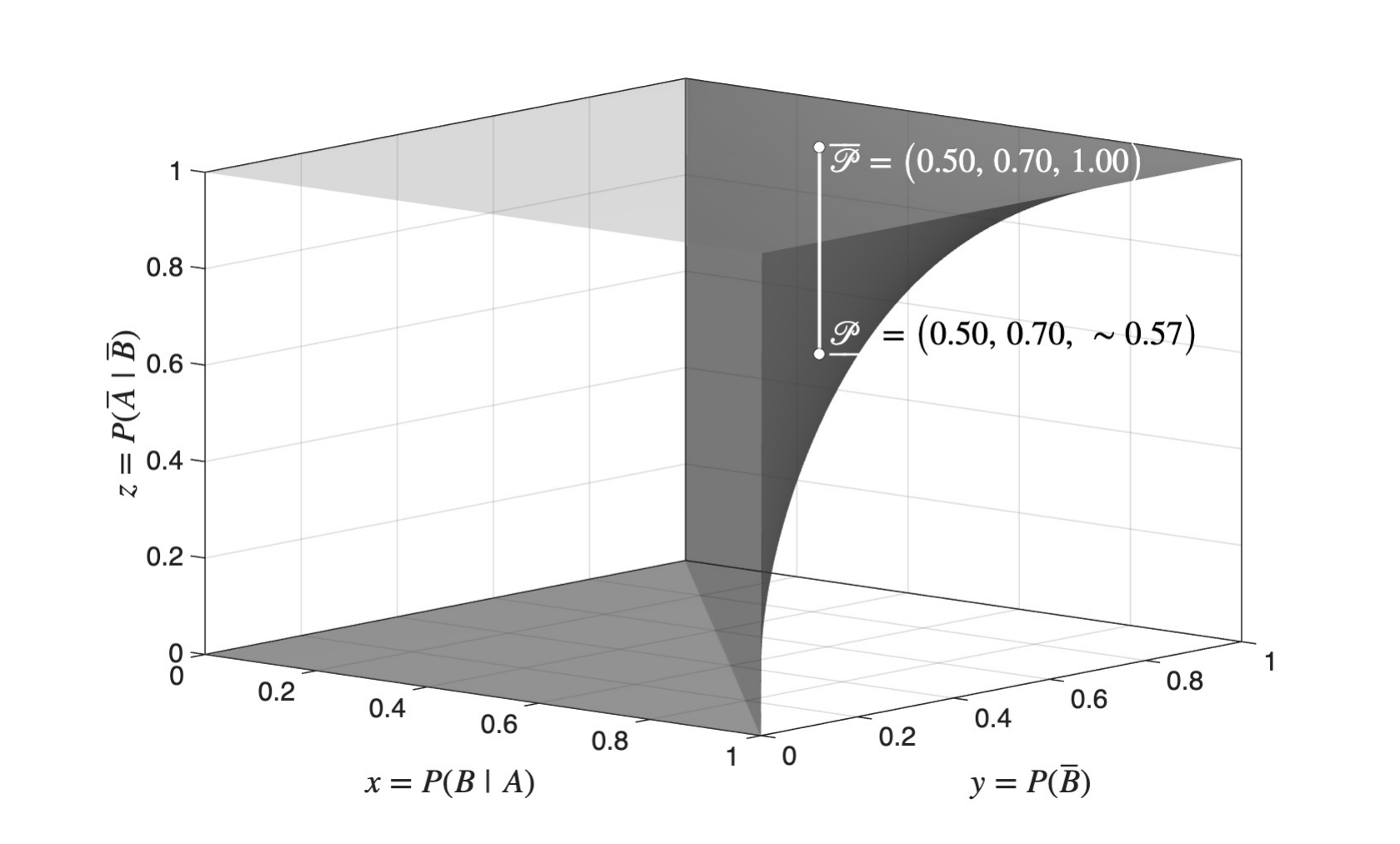}
\caption{Geometric representation of the  set 
$\Pi$ of all coherent probability assessments $(x,y,z)$ on the family 
$\mathcal{F}=\{B|A,\; \no{B},\; \no{A}|\no{B}\}$ (likewise on $\mathcal{F}=\{B|AH,\; \no{B}|H,\; \no{A}|\no{B}H\}$) associated with Weak Contraposition. 
For each $(x,y)\in[0,1]^2$, the admissible values of $z$ lie in the interval 
$[z'(x,y),z''(x,y)]$, where
$
z'(x,y)=\max\!\left\{0,\frac{x+y-1}{xy}\right\}, \;\; z''(x,y)=1.
$
The lower surface represents the coherence bound $z'(x,y)$, while the upper plane corresponds to $z''(x,y)=1$. 
The vertical segment connects the extreme coherent assessments 
$\underline{\mathcal{P}}=(x,y,z'(x,y))$ and $\overline{\mathcal{P}}=(x,y,z''(x,y))$ at $(x,y)=(0.5,0.7)$. 
}
            \label{FIG:PiWC}
\vspace{-1cm}           
        \end{figure}

By  Theorem~\ref{THM:WC1} and Theorem~\ref{THM:WC} (and by Corollary \ref{COR:WC1} and Corollary \ref{COR:WC}) we obtain the same probability propagation rules for Weak Contraposition with and without background knowledge $H$.  For a graphical illustration of the set $\Pi$ see Figure \ref{FIG:PiWC}. Example assessments for both versions of  Weak Contraposition are given in Table \ref{TAB:ValuesWC}.
\begin{remark}
Analogous to Remark~\ref{REM:InstantiationWC}, by instantiating $x=1$ and $y>0$ in Theorem~\ref{THM:WC}, it follows that $z'(1,y)=z''(1,y)=1$, which validates  Equation~(\ref{EQ:WCwithH}). 
\end{remark}
\begin{table}[]
\centering
\begin{subtable}[t]{0.48\textwidth}
\centering
\scriptsize
\begin{tabular}{cc|c|c}
$x$ & $y$ & $z'$ & $z''$ \\
\hline
~0.00~ & ~0.00~ & ~0.00~ & ~1.00~ \\
~0.00~ & ~1.00~ & ~0.00~ & ~1.00~ \\
~0.20~ & ~0.80~ & ~0.00~ & ~1.00~ \\
~0.20~ & ~0.90~ & ~0.56~ & ~1.00~ \\
~0.20~ & ~1.00~ & ~1.00~ & ~1.00~ \\
~0.30~ & ~0.80~ & ~0.42~ & ~1.00~ \\
~0.30~ & ~0.90~ & ~0.74~ & ~1.00~ \\
~0.30~ & ~1.00~ & ~1.00~ & ~1.00~ \\
~0.40~ & ~0.70~ & ~0.36~ & ~1.00~ \\
~0.40~ & ~0.80~ & ~0.63~ & ~1.00~ \\
~0.40~ & ~0.90~ & ~0.83~ & ~1.00~ \\
~0.40~ & ~1.00~ & ~1.00~ & ~1.00~ \\
~0.50~ & ~0.60~ & ~0.33~ & ~1.00~ \\
~0.50~ & ~0.90~ & ~0.89~ & ~1.00~ \\
~0.50~ & ~1.00~ & ~1.00~ & ~1.00~ \\
\\
\end{tabular}
\end{subtable}
\hfill
\begin{subtable}[t]{0.48\textwidth}
\centering
\scriptsize
\begin{tabular}{cc|c|c}
$x$ & $y$ & $z'$ & $z''$ \\
\hline
~0.60~ & ~0.50~ & ~0.33~ & ~1.00~ \\
~0.60~ & ~0.90~ & ~0.93~ & ~1.00~ \\
~0.60~ & ~1.00~ & ~1.00~ & ~1.00~ \\
~0.70~ & ~0.40~ & ~0.36~ & ~1.00~ \\
~0.70~ & ~0.90~ & ~0.95~ & ~1.00~ \\
~0.70~ & ~1.00~ & ~1.00~ & ~1.00~ \\
~0.80~ & ~0.90~ & ~0.97~ & ~1.00~ \\
~0.80~ & ~1.00~ & ~1.00~ & ~1.00~ \\
~0.90~ & ~0.20~ & ~0.56~ & ~1.00~ \\
~0.90~ & ~0.90~ & ~0.99~ & ~1.00~ \\
~0.90~ & ~1.00~ & ~1.00~ & ~1.00~ \\
~0.99~ & ~0.02~ & ~0.51~ & ~1.00~ \\
~1.00~ & ~0.00~ & ~0.00~ & ~1.00~ \\
~1.00~ & ~0.10~ & ~1.00~ & ~1.00~ \\
~1.00~ & ~1.00~ & ~1.00~ & ~1.00~ \\
\\ 
\end{tabular}
\end{subtable}
\caption{Example instantiations of probabilistic Weak Contraposition (Theorems~\ref{THM:WC1} and ~\ref{THM:WC}). 
Here $x=P(B|A)$ (resp., $x=P(B|AH)$) and $y=P(\no{B})$ (resp., $x=P(\no{B}|H)$). The table reports the corresponding coherent bounds 
$z'$ and $z''$ for $z=P(\no{A}|\no{B})$ (resp., $z=P(\no{A}|\no{B}H)$). Note that the probability propagation rules for Weak 
Contraposition with and without the background knowledge $H$ coincide.}
    \label{TAB:ValuesWC}
\vspace{-0.7cm}
\end{table}
We now turn to the interval-valued case of Weak Contraposition, where the premises are assessed by probability intervals.
\begin{theorem}\label{THM:IWC}
  Let $A,B,H$ be any logically independent events,  with $A,\no{B},H \neq \emptyset$, and  let $\mathcal{A}=[x_1,x_2]\times[y_1,y_2]\subseteq [0,1]^2$ be an interval-valued probability assessment on $\{B|AH,\no{B}|H\}$. Then, the interval of coherent extensions of the interval-valued probability assessment $\mathcal{A}$ to $\no{A}|\no{B}H$  is the interval $[z^*(\mathcal{A}),z^{**}(\mathcal{A})]$ where 
\[z^*(\mathcal{A})=\left\{
 \begin{array}{ll}
 \frac{x_1+y_1-1}{x_1y_1}, & \text{ if } x_1+y_1>1,\\
 0, &\text{ if } x_1+y_1\leq 1 \, .
 \end{array}
 \right.\quad
\text{ and }\quad z^{**}(\mathcal{A})=1.
\] 
 \end{theorem}
\begin{proof}
    The proof is straightforward by recalling that every assessment $(x,y)\in[0,1]^2$ on $\{B|AH,\no{B}|H\}$ is coherent  and by observing that  both the lower and upper bounds $z'(x,y)$ and $z''(x,y)$ for $\no{A}|\no{B}H$ given in Theorem \ref{THM:WC}  are non-decreasing  functions in the arguments  $x$ and $y$. 
\end{proof}
Notice that the interval of coherent extensions $[z^*(\mathcal{A}),z^{**}(\mathcal{A})]$ in Theorem \ref{THM:IWC} coincides in the case where $H=\Omega$.
\section{Weak Contraposition and Iterated Conditionals}\label{SEC:ITERATED}
We recall that in the theory of logical operations among conditional events in the setting of conditional random quantities (see, e.g., \cite{CaSa24,GiSa14,GOPSai24,SPOG18}), iterated conditionals can be managed without running into Lewis' triviality results (\cite{lewis76}). Moreover, in this framework all basic logical and  probabilistic properties are preserved.

We recall the notions of p-consistency and p-entailment for  conditional random quantities (and in particular for conditional events)  in the setting of coherence (\cite{SPOG18}).
Let $\mathcal{F}_n = \{X_i|H_i \, , \; i=1,\ldots,n\}$ be  a family of $n$  conditional random quantities which take values in a finite subset of $[0,1]$. Then, $\mathcal{F}_n$ is  {\em p-consistent} if and only if,
	the (prevision) assessment $(\mu_1,\mu_2,\ldots,\mu_n)=(1,1,\ldots,1)$ on $\mathcal{F}_n$ is coherent. Moreover,
	a p-consistent family  $\mathcal{F}_n$ {\em p-entails} a conditional random quantity $X|H\in[0,1]$, denoted by $\mathcal{F}_n \; \Rightarrow_p \; X|H$, if and only if  for any  coherent (prevision) assessment $(\mu_1,\ldots,\mu_n,z)$ on $\mathcal{F}_n \cup \{X|H\}$: if $\mu_1=\cdots=\mu_n=1$, then  $z=1$. The inference from   $\F_n$ to $X|H$ is \emph{p-valid} if and only if  $\mathcal{F}_n \; \Rightarrow_p X|H$.
\begin{remark}\label{REM:PvalWC}
By instantiating $x=1$ and $y=1$ in Theorem~\ref{THM:WC1}, we observe that $z'(1,1)=z''(1,1)=1$, which validates the p-validity of Weak Contraposition (for the notion of p-validity in the setting of coherence, see, e.g., \cite{GOPSai24}). That is
$
     \{B|A,\no {B}\} \Longrightarrow_p \no A|\no{B}.
$
Moreover, by \cite[Theorem 8]{GiPS20} it follows that 
$
(\no{A}|\no{B})|((B|A) \wedge \no {B})=1,
$
that is the conditional random quantity $(\no{A}|\no{B})|((B|A) \wedge \no {B})$ is constant and equal to 1. Likewise, for the case of Weak Contraposition with knowledge base $H$ we obtain that
\[
(\no{A}|\no{B}H)|((B|AH) \wedge (\no {B}|H))=1.
\]
\end{remark}
We now introduce a  version of Weak Contraposition for iterated conditionals, where events are replaced by conditional events, and probability $P$ is replaced by prevision $\prev$ for conditional random quantities (which we use to represent iterated conditionals). Formally the resulting rule is: 
\begin{equation}\label{EQ:IteratedWC}
\prev[(B|K)|(A|H)]=1,\; P( \no{B}|K)>0 
\Longrightarrow \prev[(\no{A}|H)|(\no{B}|K)]=1,
\end{equation}
or equivalently
\begin{equation}\label{EQ:IteratedWCBIS}
\prev[(B|K)|(A|H)]=1,\; P( B|K)<1 
\Longrightarrow \prev[(\no{A}|H)|(\no{B}|K)]=1,
\end{equation}

In the next result we give the probability propagation rule for  Weak Contraposition for iterated conditionals.
\begin{theorem}\label{THM:IteratedWC}
	{\rm
 Given the family $\{(B|K)|(A|H), \no{B}|K\}$, with $A,B,H, K$ logically independent, the prevision assessment $\nu = \prev[(\no{A}|H)|(\no{B}|K)]$ is a coherent extension of $\prev[(B|K)|(A|H)]=\mu\in[0,1]$ and $P(\no{B}|K)=y\in[0,1]$ if and only if  $\nu\in [\nu'(\mu,y),\nu''(\mu,y)]$, where 
 \begin{equation}
     \label{EQ:LB&UBITER} 
  \nu'(\mu,y)=\left\{
 \begin{array}{ll}
 \frac{\mu+y-1}{\mu y}, & \text{ if } \mu+y>1,\\
 0, &\text{ if } \mu+y\leq 1 \, .
 \end{array}
 \right.\quad
\text{ and }\quad \nu''(\mu,y)=1.
\end{equation} 
}
\end{theorem}
\begin{proof}
   The proof is omitted due to lack of space. 
\end{proof}
\begin{corollary}\label{COR:WCITER}
Let $A,B,H,K$ be any logically independent events. Then, the set $\Pi$ of all coherent prevision assessments $(\mu,y,\nu)$ on the family $\F=\{(B|K)|(A|H), \no{B}|K,(\no{A}|H)|(\no{B}|K) \}$, is 
$\Pi=\{(\mu,y,\nu): (\mu, y)\in[0,1]^2, \nu \in [\nu'(\mu,y), \nu''(\mu,y)]\},	
$
 where  $\nu'(\mu,y)$ and $\nu''(\mu,y)$ are  defined in (\ref{EQ:LB&UBITER}).
\end{corollary}
\begin{remark}
The 
set $\Pi$ in Corollary \ref{COR:WCITER} coincides with the set $\Pi$ in Corollary \ref{COR:WC1}, where $x$ and $z$ are replaced by $\mu$ and $\nu$, accordingly. In this sense, Figure \ref{FIG:PiWC} is also a geometric representation of the set $\Pi$ in Corollary  \ref{COR:WCITER} when events are replaced by conditional events.
\end{remark}
\begin{remark}
    By instantiating $\mu=1$ and $y>0$ in Theorem~\ref{THM:IteratedWC}, it follows that $\nu'(1,y)=\nu''(1,y)=1$, which validates  rules~(\ref{EQ:IteratedWC}) and (\ref{EQ:IteratedWCBIS}).   If  $\mu=y=1$ then $
\{(B|K)|(A|H), \no{B}|K) \}
\Longrightarrow_p (\no{A}|H)|(\no{B}|K).
$

\end{remark}
For an everyday  sample of Weak Contraposition for iterated conditionals, where 
the two premises 
(P1) $\prev[(B|K)|(A|H)]=1$ and (P2) $P(B|K)<1$ imply 
(C)~$\prev[(\no{A}|H)|(\no{B}|K)]=1$, consider:
\\
{\small
(P1) If $\overbrace{\text{\emph{Lorenzo studies hard when he wants to pass the exam}}}^{A|H}$, then 

  normally \emph{$\overbrace{\text{he will  pass the exam when he takes it}}^{B|K}$}.     \\ 
\noindent (P2)
It is not the case that: normally $\overbrace{\text{\emph{Lorenzo will  pass the exam when he takes it}}}^{B|K}$. \\ 
(C) If $\overbrace{\text{\emph{Lorenzo will  not pass the exam when he takes it}}}^{\no B |K}$, then

normally $\overbrace{\text{\emph{Lorenzo does not study hard when he wants to pass the exam}}}^{\no A | H}$. 
}
\section{Concluding Remarks}
\label{SEC:ConcludRem}
In this paper, we have investigated the notion of Weak Contraposition in a probabilistic setting, providing both theoretical foundations and practical probability propagation rules. We showed that, unlike classical Contraposition, Weak Contraposition preserves nonmonotonicity and can be given a coherent probabilistic semantics. In particular, we derived explicit coherent bounds for the probability of the conclusion (with and without background knowledge) and extended the framework to iterated conditionals, demonstrating that the rules remain preserved when  events are replaced by conditional events. We also studied Weak Contraposition involving interval-valued assessments and offering corresponding  probability propagation rules that naturally generalize the point-valued case. 

From a cognitive perspective, Weak Contraposition aligns better with intuitive reasoning patterns, as experimental evidence suggests that people rarely endorse  Contraposition in everyday reasoning. By formalizing a probabilistically coherent version of this inference scheme, we provide a bridge between normative probabilistic reasoning and descriptive models of human inference.

Future work includes empirical studies to assess how people process and endorse probabilistic Weak Contraposition, as well as further formal generalizations. Overall, our study contributes to a deeper understanding of conditional reasoning under uncertainty, highlighting how subtle modifications to classical rules can yield both logically sound and psychologically plausible inference patterns. \\       
\textbf{Acknowledgements.}
{\small Niki Pfeifer was supported by European Research Council (ERC) under the European Union’s Horizon 2020 [ERC-2020-ADG, 101018280, ConLog]. Giuseppe Sanfilippo was supported by the FFR project of University of Palermo and by INdAM-GNAMPA research group.}
{\small
\bibliographystyle{spbasic}

}
%
%

\end{document}